\documentclass[11pt,a4paper]{article}
\usepackage{times,latexsym}
\usepackage{url}
\usepackage[T1]{fontenc}
\usepackage[acceptedWithA]{tacl2021v1}
\usepackage{amsmath}
\usepackage{graphicx}
\usepackage{booktabs}
\usepackage{microtype}
\hypersetup{hidelinks}

\makeatletter
\newcommand{\tablecaption}[1]{\def\@captype{table}\caption{#1}}
\newcommand{\figurecaption}[1]{\def\@captype{figure}\caption{#1}}
\makeatother

\title{The Depth of a Prediction: Where Language Models Resolve a Word Predicts Human Reading Time}
\author{Donato Riccio \\
  Department of Engineering and Science \\
  Universitas Mercatorum, Rome, Italy \\
  \texttt{donato.riccio@studenti.unimercatorum.it}
  \AND
  Fabrizio Maturo \\
  Department of Economics, Statistics and Business \\
  Universitas Mercatorum, Rome, Italy \\
  \texttt{fabrizio.maturo@unimercatorum.it}}

\begin{document}
\maketitle

\begin{abstract}
Surprisal from a language model's final layer predicts how long people read a word, yet the model computes its prediction across many layers.
We read each word's prediction at every layer with the tuned lens and summarize this layerwise surprisal trajectory by its resolution depth, the depth around which its change is centered.
On eight English eye-tracking and self-paced reading corpora (81,957 words) and ten GPT-2, OPT and Pythia models, words that a model resolves later, and the words after them, are read more slowly than equally surprising words.
Resolution depth improves held-out prediction beyond final-layer surprisal, entropy and the best single layer by 0.26 millinats per word (pooled $p = 0.001$), about a third of entropy's gain, and each model family gains, GPT-2 and Pythia through one corpus each.
The net gain varies between corpora (random-effects mean 0.07, $p = 0.37$); against layer-shuffled depth, which adds as many coefficients, it is positive in seven of eight corpora (random-effects mean 0.23, $p = 0.016$), and untrained networks show no gain under the logit lens.
Pooled over corpora, the effect holds with corrected word probabilities and beyond first-to-last-layer divergences, and a whole-trajectory functional regression adds no significant information beyond resolution depth.
\end{abstract}

\section{Introduction}

Readers spend longer on words that are hard to predict.
Surprisal theory makes this precise: the processing cost of a word grows with its surprisal, the negative log probability of the word given its context \citep{hale2001probabilistic,levy2008expectationbased}.
Neural language models supply these probabilities, and their surprisal predicts reading times on held-out data across languages, corpora and reading paradigms \citep{wilcox2023testing,shain2024largescale}.
In these studies the probability comes from the model's output layer.

A transformer language model reaches that output in steps.
Each block updates a residual stream, and decoding the stream after every block, with the model's own unembedding (the logit lens; \citealp{nostalgebraist2020interpreting}) or with a learned affine translator for each layer (the tuned lens; \citealp{belrose2023eliciting}), yields a sequence of predictions that converges toward the output.
These intermediate predictions bear on human reading.
Surprisal read from intermediate layers fits self-paced reading times and gaze durations better than final-layer surprisal \citep{kuribayashi2025large}; the probability update between the first and the last layer adds to surprisal in explaining reading times for garden-path sentences \citep{kuribayashi2026dual}; and in recall, categorization and reasoning tasks, measures of how a forward pass unfolds, such as the layer after which the model settles on an answer, improve predictions of human response times beyond final-layer measures \citep{hu2025signatures}.

The reading-time studies treat depth either as a choice of layer for a whole dataset or as a contrast between two layers.
Within one model, however, the prediction of each word follows its own course across depth: some predictions are fixed in the first blocks while others change until the last \citep{geva2022transformer,belrose2023eliciting}.
Whether the position along depth at which a word's prediction resolves carries information about reading effort, beyond how surprising the word finally is, has not been tested.
A test needs a per-word measure of timing that is separate from the size of the change, a comparison with the best single layer, and held-out evaluation, because layerwise predictors are numerous and strongly correlated.

\begin{figure}[t]
\centering
\includegraphics[width=0.86\columnwidth]{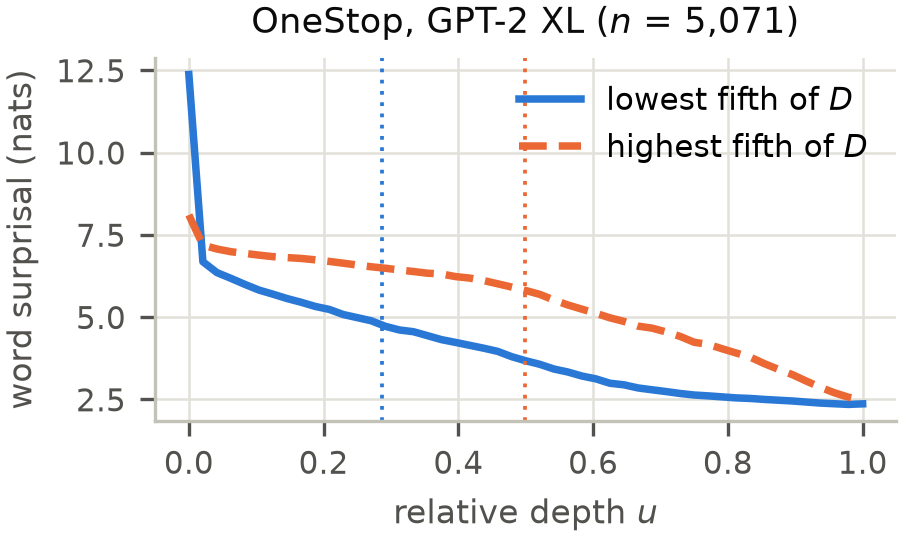}
\caption{Mean layerwise surprisal trajectories (tuned lens) of OneStop words in the middle fifth of final surprisal (40th to 60th percentile) under GPT-2 XL, split into the fifth with the lowest and the fifth with the highest resolution depth $D$ (Natural Stories with Pythia 2.8B in Figure~\ref{fig:appendix}a). Dotted lines mark each group's mean $D$; $n$ is the number of analyzed words in the band. Both groups end at the same surprisal; the late-resolved words keep a high surprisal through more of the network.}
\label{fig:trajectories}
\end{figure}

We treat each word's surprisal across layers as a function of relative depth, its layerwise surprisal trajectory (Figure~\ref{fig:trajectories}), and summarize where the trajectory changes by its resolution depth $D$: the centroid of the trajectory's absolute change along relative depth.
$D$ lies between 0 and 1, equals 0.5 when the change is spread evenly over the layers whatever their number, and is defined separately from the final surprisal and the total amount of change, both of which enter our analyses as controls.
We test $D$ on eight English reading corpora with ten models from three families, against controls that include final-layer surprisal and entropy, frequency, length, bigram predictability, spillover, the amount of layerwise change and the best single layer, using cross-validated log-likelihood and permutation tests pooled over corpora and models, and test with a functional regression on the whole trajectory whether $D$ summarizes its predictive information.

Words that a model resolves later, and the words that follow them, are read more slowly than equally surprising words.
Pooled over corpora and models, $D$ improves held-out log-likelihood by 0.26 millinats per word (95\% CI 0.10 to 0.40; $p = 0.001$), about a third of the gain from contextual entropy, and each of the three families gains on its own: GPT-2 through OneStop, Pythia through Natural Stories and OPT without either.
The net gain varies between corpora: in the smallest corpora two added coefficients cost more held-out likelihood than $D$ supplies, as they do for layer-shuffled and untrained $D$, and its random-effects mean is 0.07 ($p = 0.37$).
Against layer-shuffled $D$, which adds as many coefficients, the gain is positive in seven of the eight corpora, and its random-effects mean of 0.23 (95\% CI 0.03 to 0.44; $p = 0.016$) shows that, across corpora, $D$ carries information beyond the held-out cost of its two coefficients.
Untrained networks read with the logit lens do not reproduce the gain, and it survives corrected word probabilities, the removal of the embedding readout, lexical, skipping and endpoint controls and first-to-last-layer divergences.

The paper thus contributes a per-word measure of where along a transformer's depth the prediction of a word resolves (\S\ref{sec:depth}), evidence from eight corpora and ten models of three families that this measure predicts held-out reading times beyond surprisal and the best single layer and loses its gain when the layers are shuffled (\S\ref{sec:main}--\S\ref{sec:robust}), a functional analysis showing that, pooled over corpora, it summarizes the trajectory's predictive information, while OneStop and Natural Stories hold more in the trajectory's shape (\S\ref{sec:functional}, \S\ref{sec:res-functional}), and a characterization of the words that models resolve late (\S\ref{sec:words}).

\section{Related Work}

\paragraph{Surprisal and reading times.}
Surprisal links probabilistic prediction to processing cost \citep{hale2001probabilistic,levy2008expectationbased}.
Language-model surprisal predicts held-out reading times in eleven languages \citep{wilcox2023testing} and, across more than 2.2 million data points from six English datasets, with a cost linear in surprisal \citep{shain2024largescale}; contextual entropy adds to surprisal in three of four datasets \citep{pimentel2023effect}; and within a model family, larger models give surprisal that fits reading times worse \citep{oh2023why}.
Other model-internal quantities also predict reading times: hidden-state representations of early layers predict first-pass measures better than surprisal \citep{tsipidi2026probing}, and the discretization timestep of a Mamba state-space model \citep{yamamoto2026timesteps}, the energy of an energy-based transformer \citep{dotlacil2026energy} and deviations from the path of hidden states across successive words \citep{barenholtz2026trajectory} add to surprisal.
Vector predictors from larger models gain largely through added dimensions, and untrained networks of the same size reproduce much of their fit, which makes untrained models a necessary baseline \citep{lin2026vectors}.

\paragraph{Model layers and human processing.}
\citet{kuribayashi2025large} read surprisal from every layer of 30 models with the logit lens, and of 14 of them with the tuned lens, and found that intermediate layers fit self-paced reading and first-pass gaze durations better than the final layer, with faster measures aligned to earlier layers; the analysis selects the best layer per dataset and measure and evaluates fit in sample.
\citet{kuribayashi2026dual} report the reverse alignment in syntactically challenging regions and show that probability updates between the first and the last layer (a surprisal difference and the Kullback--Leibler and Jensen--Shannon divergences) improve fits to self-paced reading of garden-path sentences.
\citet{hayakawa2026reader} summarize where across layers the predictive power of surprisal concentrates, as a mean layer weighted by log-likelihood gains, for groups of second-language readers and gaze measures.
\citet{hu2025signatures} relate the time course of a forward pass to human response times and accuracy in decision tasks without reading data.
Resolution depth locates each word's own prediction change along depth and enters the regression as a word-level predictor, evaluated on held-out naturalistic reading against the best single layer and against first-to-last-layer measures.

\paragraph{Depth of computation.}
The depth at which a network's prediction becomes fixed varies across inputs and indexes example difficulty in image classifiers \citep{baldock2021deep}; the tuned lens defines an analogous prediction depth for language-model tokens \citep{belrose2023eliciting}; and the layers at which the top-ranked tokens saturate vary across tokens \citep{geva2022transformer,lioubashevski2025looking}, with ordered saturation present even in untrained networks \citep{lioubashevski2025looking}.
These indices track the stability of the top-ranked token against model-internal quantities; resolution depth is a continuous centroid of the surprisal change of the word that occurs, validated against human reading times.

\section{Resolution Depth}
\label{sec:depth}

\subsection{Layerwise surprisal trajectories}

Consider a decoder-only transformer with $L$ blocks and hidden states $h_0, \dots, h_L$ at each position, where $h_0$ is the embedding output and $h_i$ the residual stream after block $i$.
A readout maps $h_i$ to a distribution over the next token.
The logit lens applies the model's final normalization and unembedding directly, $p_i = \mathrm{softmax}(W_U \, \mathrm{LN}(h_i))$ \citep{nostalgebraist2020interpreting}.
The tuned lens first applies a learned affine translator, $p_i = \mathrm{softmax}(W_U \, \mathrm{LN}(h_i + A_i h_i + b_i))$, trained to match the model's output distribution \citep{belrose2023eliciting}.
The readout at $i = L$ is the model's output.
For a word $w$ that spans tokens $t_1, \dots, t_k$ in context $c$, its surprisal at readout $i$ is
\begin{equation}
s_i(w) = -\sum_{j=1}^{k} \log p_i(t_j \mid c, t_{<j}),
\end{equation}
and readout $i$ sits at relative depth $i/L$.
The sequence $s_0(w), \dots, s_L(w)$ is the word's layerwise surprisal trajectory (Figure~\ref{fig:trajectories}).

\subsection{Definition}

The change between readouts $i-1$ and $i$ spans the depths from $(i-1)/L$ to $i/L$, and we place it at the midpoint $v_i = (i - \tfrac{1}{2})/L$.
The resolution depth of $w$ is the centroid of the trajectory's absolute change along relative depth:
\begin{equation}
D(w) = \frac{\sum_{i=1}^{L} v_i \, \lvert s_i(w) - s_{i-1}(w) \rvert}{\sum_{i=1}^{L} \lvert s_i(w) - s_{i-1}(w) \rvert}.
\label{eq:depth}
\end{equation}
$D$ is a weighted mean of the step depths $v_1, \dots, v_L$, with weights proportional to how much the word's surprisal changes at each layer step.
A word whose surprisal falls in the first blocks and then stays flat has a low $D$; a word whose surprisal keeps changing until the top has a high $D$; a change spread evenly over the layers gives $D = 0.5$ for any number of layers.
Writing $\mathrm{TV}(w) = \sum_{i} \lvert s_i(w) - s_{i-1}(w) \rvert$ for the total variation of the trajectory, which is positive for every word we analyze, the resolution profile $\pi_i(w) = \lvert s_i(w) - s_{i-1}(w) \rvert / \mathrm{TV}(w)$ is a distribution over depth, and $D(w) = \sum_i v_i \pi_i(w)$ is its mean.
For a trajectory that decreases monotonically, $D(w) = \bigl(\bar{s}(w) - s_L(w)\bigr) / \bigl(s_0(w) - s_L(w)\bigr)$, where $\bar{s}(w)$ is the mean of the trajectory over relative depth (trapezoidal rule on the readouts): $D$ is the area between the trajectory and its final value as a fraction of its total decrease, so a word with high $D$ keeps a high surprisal through more of the network.
It describes where the trajectory changes; the final surprisal $s_L(w)$, the total variation $\mathrm{TV}(w)$ and the surprisal update $\mathrm{SU}(w) = s_0(w) - s_L(w)$ describe how much, and they enter our analyses as controls.

We use the centroid because a threshold landmark, the first depth at which the trajectory completes 90\% of its cumulative change, compresses into the top of the network: tuned-lens trajectories keep changing until the output, and across our ten models the landmark falls in the top fifth of depth for 60\% to 89\% of words.
The centroid spreads over the middle of the network (median 0.38 to 0.50 by model).

\section{Data and Models}
\label{sec:data}

\subsection{Reading corpora}

We use eight publicly released English reading corpora (Table~\ref{tab:corpora}).
In Provo \citep{luke2018provo}, 84 native readers of American English read 55 short passages from news, popular-science and fiction sources.
In GECO \citep{cop2017presenting}, 14 English monolinguals read a complete novel, Agatha Christie's \textit{The Mysterious Affair at Styles}, in screens of up to 145 words.
From OneStop \citep{berzak2025onestop} we use the ordinary-reading regime, in which 180 native readers each read ten of 30 Guardian articles, paragraph by paragraph, in an original or a simplified version; we exclude practice and repeated-reading trials.
From MECO \citep{siegelman2022expanding} we use the English sample of Wave~1 (46 readers) in release 2.0, version 2.1 \citep{siegelman2025meco}, who read 12 encyclopedic texts based on Wikipedia entries.
From ZuCo 1.0 \citep{hollenstein2018zuco} we use the normal-reading task on Wikipedia sentences, read one sentence at a time by 12 native speakers.
In Natural Stories \citep{futrell2021natural}, 181 native speakers read ten edited texts, two of them adapted from Wikipedia articles, word by word in a moving-window self-paced reading paradigm.
The UCL corpus \citep{frank2013reading} contributes one corpus of each kind: 361 sentences from novels, chosen to be interpretable out of context and read one word at a time, and the 205 that fit on one line, read on an eye tracker by other participants.

\begin{table*}[t]
\centering
\small
\setlength{\tabcolsep}{5pt}
\begin{tabular}{llrrrrr}
\toprule
Corpus & Material & Readers & Words & Analyzed & Units & Readers/word \\
\midrule
Provo & 55 passages & 84 & 2,745 & 2,115 & 55 passages & 49 \\
GECO & one novel & 14 & 54,362 & 33,624 & 586 screens & 7 \\
OneStop & 30 articles, 2 versions & 180 & 35,188 & 25,359 & 30 articles & 14 \\
MECO & 12 texts & 46 & 2,107 & 1,732 & 12 texts & 18 \\
ZuCo & 300 sentences & 12 & 6,588 & 4,711 & 300 sentences & 9 \\
UCL eye tracking & 205 sentences & 37 & 1,931 & 1,305 & 205 sentences & 25 \\
Natural Stories & 10 stories & 181 & 10,256 & 9,237 & 10 stories & 84 \\
UCL self-paced & 361 sentences & 62 & 4,957 & 3,874 & 361 sentences & 39 \\
\midrule
Total & & & 118,134 & 81,957 & 1,559 units & \\
\bottomrule
\end{tabular}
\caption{Reading corpora: six eye-tracking and two self-paced reading corpora (Natural Stories, UCL self-paced); the two UCL corpora present the same sentences (205 in eye tracking) to different readers. Words are the displayed word tokens of each release; analyzed words enter the primary analysis (go-past time for eye tracking, reading time for self-paced reading) after exclusions. Units define the cross-validation folds and the units of inference (\S\ref{sec:analysis}). Readers/word is the median number of readers contributing to an analyzed word. Total reading time analyzes 36,382 words in GECO, 26,824 in OneStop, 1,881 in MECO and 1,316 in UCL eye tracking, and the words of the primary measure elsewhere.}
\label{tab:corpora}
\end{table*}

For eye tracking we use first fixation duration, gaze duration (first-pass dwell time), go-past time (regression-path duration) and total reading time.
The first three count readers who fixated the word during first-pass reading; total reading time counts every reader who fixated the word.
For each word and measure we average log durations over the contributing readers and keep words with at least five contributing readers (three in GECO, which has 14 readers in total).
Self-paced reading contributes one measure, the log reading time.
We exclude words at sentence and line edges, words without letters and words whose labels disagree between text and reading data; Appendix~\ref{app:impl} gives the details and the reader criteria of Natural Stories and UCL.
Go-past time is the primary eye-tracking measure and reading time the measure for self-paced reading; we report all four eye-tracking measures.
Go-past time is the latest first-pass measure: it includes the regressions a word triggers before the reader moves past it, and surprisal from later layers aligns with slower measures of processing \citep{kuribayashi2025large}, so a measure of where the prediction resolves along depth is most directly tested on it.

\subsection{Models and readouts}

We use GPT-2 small, large and XL \citep{radford2019language}, whose released checkpoints have 124M, 774M and 1.5B parameters and 12, 36 and 48 blocks; OPT 125M, 1.3B and 6.7B \citep{zhang2022opt}, with 12, 24 and 32 blocks; and Pythia-deduped 410M, 1.4B, 2.8B and 6.9B \citep{biderman2023pythia}, with 24, 24, 32 and 32 blocks.
These are the three families for which \citet{kuribayashi2025large} also used the released tuned lenses \citep{belrose2023eliciting}.
The families differ in training data: GPT-2 was trained on WebText, from which Wikipedia documents were removed \citep{radford2019language}, and OPT and Pythia on data that include English Wikipedia through the Pile \citep{gao2020pile,zhang2022opt,biderman2023pythia}.
As untrained controls we read four networks with the logit lens, since tuned lenses exist only for trained models: GPT-2 XL and OPT 1.3B with random weights, since these families release no initial checkpoint, and Pythia 410M and 1.4B at their released initialization checkpoints.

\section{Analysis}
\label{sec:analysis}

\subsection{Regression and controls}

For each corpus, model and measure we predict the mean log reading time $y_w$ of word $w$ by weighted least squares with weights equal to the number of contributing readers $n_w$, which corresponds to a Gaussian likelihood with variance $\sigma^2 / n_w$.
The baseline contains word length, Zipf frequency \citep{speer2022wordfreq} and their product for $w$, $w-1$ and $w-2$, the log position of $w$ in its sentence, and the surprisal of $w$ and $w-1$ under a word bigram model of WikiText-103 \citep{merity2017pointer} with interpolated absolute discounting.
The control set adds the language-model quantities: final-layer surprisal of $w$ and its square, final-layer surprisal of $w-1$ and $w-2$, final-layer entropy of $w$ and $w-1$, the total variation and surprisal update of $w$, the log number of tokens of $w$ and $w-1$, and the surprisal of $w$ and $w-1$ at the best single non-final readout.
The best readout is chosen inside each training fold by five-fold grouped cross-validation over readouts $0, \dots, L-1$, so that the control set contains the strongest single-layer predictor available to the model.
The test model adds $D(w)$ and $D(w-1)$ to the control set.

\subsection{Evaluation and inference}

We evaluate by ten-fold cross-validation with folds grouped by text unit (Table~\ref{tab:corpora}) and standardize predictors within training folds.
The gain of a model over its comparator is the mean per-word difference in held-out log-likelihood, reported in millinats per word.
Inference treats text units as independent: we sum the gains within each unit and test their total with a one-sided sign-flip permutation test (10,000 sign flips for one corpus, model and measure), one-sided because the hypothesis, that $D$ adds information, is directional.
To pool, we sum the unit gains of the models being pooled within each corpus and concatenate units across corpora, a sentence read in both UCL corpora forming one unit (1,354 units for the primary measure; 20,000 sign flips); the pooled gain is the summed gain divided by the number of word predictions.
Intervals are 95\% percentile intervals from 2,000 bootstrap samples of units drawn within corpora, the UCL corpora forming one.
To treat the corpora as a sample, we fit DerSimonian--Laird random-effects models to the per-corpus gains, with a Hartung--Knapp interval, the share $I^2$ of between-corpus variation beyond sampling error and a prediction interval for a new corpus, also counting the two UCL corpora as one.
To express magnitude, we compare the gain with two references on the same words and folds: final-layer surprisal of $w$ and $w-1$ added to the baseline, and final-layer entropy of $w$ and $w-1$ added to the rest of the control set.
The analyses are exploratory, and we report every corpus, model and measure we analyzed (Appendix~\ref{app:cells}).

\subsection{Layer order and untrained networks}
\label{sec:controls-def}

If $D$'s information lies in the course of the trajectory across depth, scrambling the order of readouts should remove it.
We apply five fixed random permutations to readouts $0, \dots, L-1$, keep the output last, recompute $D$ and rerun the whole cross-validated test for each, and report the median gain over the five permutations, which no single permutation determines.
Shuffled $D$ adds as many coefficients as $D$, so the difference between their gains on each text unit separates the information that $D$ draws from the trajectory from the held-out cost of the coefficients.
If the effect arose from architecture or tokenization alone, untrained networks with the same architecture and tokenizer would reproduce it; we run the complete pipeline with the four untrained networks on every corpus and measure.

\subsection{Functional regression}
\label{sec:functional}

$D$ compresses the trajectory into one number.
To test whether the whole trajectory carries more, we fit a scalar-on-function regression \citep[ch.~15]{ramsay2005functional}: the term $\int_0^1 \beta(u)\, s_w(u)\, du$, where $s_w(i/L) = s_i(w)$, discretized as $L^{-1} \sum_{i=0}^{L} \beta(i/L)\, s_i(w)$, with $\beta$ expanded in eight cubic B-splines on equally spaced knots and a second-difference roughness penalty \citep{eilers1996flexible} whose weight is chosen from $10^{-1}$ to $10^{8}$ by five-fold grouped cross-validation inside each training fold.
The model contains one such term for $w$ and one for $w-1$.
We compare the control set plus the trajectory term with the control set plus $D$, test the trajectory beyond $D$ and $D$ beyond the trajectory, and fit the trajectory term to layer-shuffled trajectories.
Because $D$ is the mean of the resolution profile $\pi(w)$, a functional regression on the profile, $\sum_{i} \beta(v_i)\, \pi_i(w)$, reproduces $D$ when $\beta$ is linear; added to the control set and $D$, it tests whether the shape of the profile carries information beyond its mean.

\subsection{Robustness and alternative measures}

We repeat the main test with $D$ computed from readouts $1, \dots, L$, leaving out the embedding readout; with word probabilities corrected for word boundaries \citep{oh2024leading,pimentel2024compute}; on words that are a single token for the model; with the function-word status of $w$ and $w-1$, the squares and cubes of their length and Zipf frequency, the squared surprisal update and the product of final surprisal and update, the total variation and surprisal update of $w-1$, or the skipping rates of $w$ and $w-1$ added to the controls; with $D(w)$ or $D(w-1)$ alone in the test model; and with $D$ computed from the entropy trajectory $H_i(w)$ of the readouts.
Under the logit lens we compute $D$ from all readouts and from readout 1, with the main controls and with total variation, surprisal update and best readout also from the logit lens.
We compare $D$ with the closest cross-layer measures: the mean surprisal $\bar{s}(w)$ of the trajectory over relative depth, also with its square and its products with final surprisal and update, and the depth of the largest single-step drop in surprisal, analogous to the area-under-the-curve and argmax-delta measures of \citet{hu2025signatures}; the Jensen--Shannon divergence and the Kullback--Leibler divergence between the readout after the first block and the output \citep{kuribayashi2026dual}, added to the controls for $w$ and $w-1$; prediction depth, the relative depth of the earliest readout from which every readout's top-ranked token equals the output's \citep{baldock2021deep,belrose2023eliciting}; and a settling depth, the first depth after which surprisal stays within 10\% of the trajectory's range from its final value.
Each alternative is tested alone and as a control for $D$.
Appendix~\ref{app:impl} gives the implementation. All corpora, models and lenses are publicly available, and we will release our analysis code under an open license with the published version.

\section{Results}
\label{sec:results}

\subsection{Resolution depth predicts reading times}
\label{sec:main}

\begin{figure}[t]
\centering
\includegraphics[width=0.86\columnwidth]{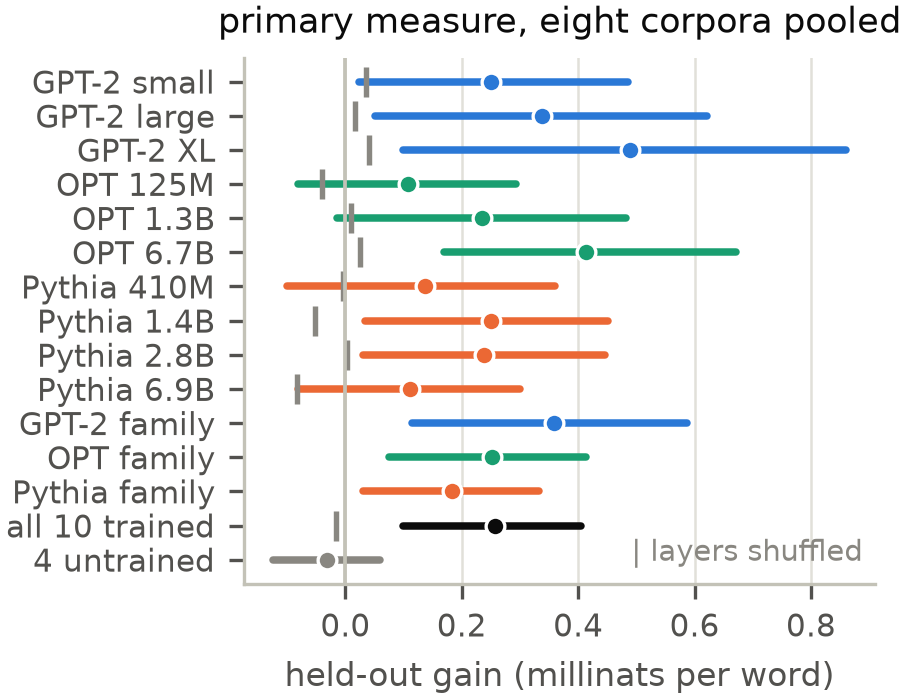}
\caption{Held-out gain in log-likelihood from adding resolution depth $D(w)$ and $D(w-1)$ to the control set on the primary measure (go-past time for eye tracking, reading time for self-paced reading), pooled over the eight corpora, per model, per family, for the ten trained models and for the four untrained networks; ticks mark the median gain over five layer permutations. Intervals are 95\% bootstrap intervals over text units drawn within corpora.}
\label{fig:main}
\end{figure}

The setup reproduces the standard surprisal effect: final-layer surprisal of $w$ improves held-out log-likelihood over the baseline in 79 of the 80 corpus--model pairs (5.94 millinats per word, pooled), and the improvement shrinks with model size within a family in 18 of 24 family--corpus series \citep{oh2023why}.
Pooled over the ten models and eight corpora, $D$ improves held-out log-likelihood on the primary measure by a further 0.26 millinats per word (95\% CI 0.10 to 0.40; $p = 0.001$; Figure~\ref{fig:main}).
Each family gains on its own: GPT-2 0.36 (0.11 to 0.59; $p = 0.004$), OPT 0.25 (0.07 to 0.41; $p = 0.002$) and Pythia 0.18 (0.03 to 0.33; $p = 0.021$).
Seven of the ten models are significant on their own, and OPT 125M, Pythia 410M and Pythia 6.9B are positive (0.11 to 0.14; $p = 0.14$ each).
On total reading time (reading time for self-paced reading), the pooled gain is 0.34 (0.15 to 0.53; $p < 0.001$).

Pooled over the six eye-tracking corpora and ten models, $D$ adds 0.09 millinats per word to first fixation duration ($p = 0.037$), 0.30 to gaze duration ($p < 0.001$), 0.21 to go-past time ($p = 0.010$) and 0.31 to total reading time ($p = 0.005$; Figure~\ref{fig:appendix}b).
GPT-2 gains on every eye-movement measure ($p \le 0.008$) and OPT on all but first fixation (0.09; $p = 0.061$); Pythia's eye-tracking gains are smaller ($-0.004$ to 0.15; $p = 0.068$ to $0.53$), and its effect concentrates in self-paced reading.

\begin{figure*}[t]
\centering
\includegraphics[width=0.85\textwidth]{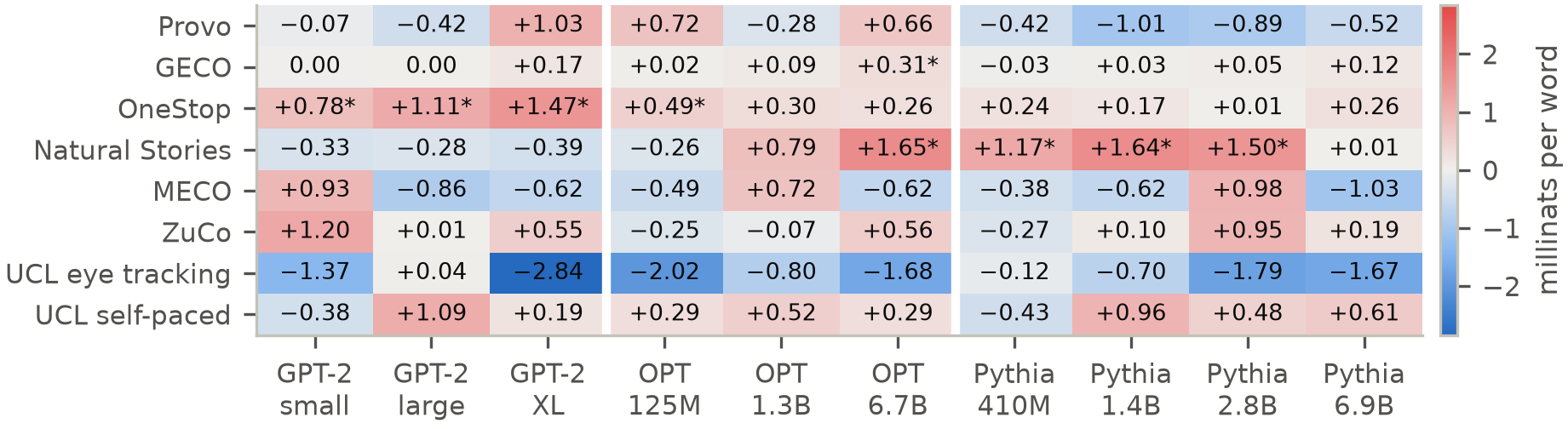}
\caption{Held-out gain from resolution depth on the primary measure for each corpus and model, in millinats per word. Asterisks mark one-sided permutation $p < 0.05$ for the corpus and model alone.}
\label{fig:map}
\end{figure*}

The effect varies across corpora (Figure~\ref{fig:map}).
It is largest for GPT-2 on OneStop (0.78 to 1.47 millinats per word; $p \le 0.009$) and on Natural Stories for Pythia 410M to 2.8B (1.17 to 1.64; $p \le 0.040$) and OPT 6.7B (1.65; $p = 0.025$).
The families differ significantly only in these two corpora: GPT-2 gains more than OPT and Pythia in OneStop (two-sided $p = 0.005$ and $0.001$), and Pythia more than GPT-2 in Natural Stories ($p = 0.008$), where GPT-2 loses ($-0.28$ to $-0.39$).
GECO gives gains near zero for every model except OPT 6.7B (0.31; $p = 0.024$); in UCL, self-paced reading gains for eight of the ten models (0.36 pooled over models; $p = 0.19$) and eye tracking loses for nine ($-1.30$); Table~\ref{tab:cells} in Appendix~\ref{app:cells} lists every corpus, model and measure.
Nine of the 80 corpus--model pairs are significant on their own (4.0 expected at $\alpha = 0.05$), and over all four measures 30 of the 260 trained cells gain significantly (13.0 expected).
Each added coefficient costs held-out likelihood, about $1/n$ nats per word for two coefficients without signal on $n$ independent words and more when the words come from few texts or short sentences: in Provo, MECO and UCL eye tracking, shuffled $D$ loses as much as $D$ (medians $-0.68$, $-0.50$ and $-1.46$ against $-0.35$, $-0.55$ and $-1.52$), so these corpora carry little information on the question.
Because the sign-flip test is centered at zero, it flags such losses more often than nominal: 30 trained cells lose significantly (two-sided $p < 0.05$; 6.5 expected in that tail), 24 of them in UCL eye tracking and Provo.
No single corpus carries the pooled effect: leaving out any one corpus keeps it significant (0.14 to 0.38 millinats per word; $p \le 0.020$).
For GPT-2 and Pythia, it rests on one corpus each: without OneStop, GPT-2 gains 0.02 ($p = 0.41$), and without Natural Stories, Pythia gains 0.07 ($p = 0.20$).
OPT keeps it without either (0.21, $p = 0.023$; 0.19, $p = 0.011$).
Treated as a sample of corpora, the net gain has a random-effects mean of 0.07 (Hartung--Knapp 95\% CI $-0.42$ to $0.56$; one-sided $p = 0.37$; $I^2 = 75\%$; 95\% prediction interval for a new corpus $-0.93$ to $1.07$), and 0.21 with the two UCL corpora counted as one ($-0.07$ to $0.48$; $p = 0.057$).
Against layer-shuffled $D$ the gain is positive in seven of the eight corpora (0.09 to 0.61; UCL eye tracking $-0.16$), with a random-effects mean of 0.23 (0.03 to 0.44; $p = 0.016$; $I^2 = 27\%$; prediction interval $-0.14$ to $0.60$; 0.25 with the two UCL corpora counted as one, $p = 0.015$), and 0.15 without OneStop and Natural Stories ($p = 0.006$); for OPT it does not vary between corpora ($I^2 = 0\%$; 0.19, 0.03 to 0.35; prediction interval 0.03 to 0.36).
GECO's few readers per word (median 7) do not explain its result: OneStop reduced to a median of three or four readers per word keeps positive gains for GPT-2 large and XL on every measure (24 of 24 draws; 12 significant).

Words resolved later, and the words after them, are read more slowly.
Added on its own, $D(w)$ gains 0.09 millinats per word ($p = 0.015$) and $D(w-1)$ 0.17 ($p = 0.002$), so much of the effect falls on the following word.
The coefficient of $D(w)$ in full-data fits is positive in 66 of the 80 pairs, and in 73\% of the 260 trained cells against 47\% of the 104 untrained cells; the coefficients of $D(w)$ and $D(w-1)$ are both positive in all nine significant pairs.
One standard deviation of $D$ lengthens go-past time by 0.6\% to 1.3\% on the word and by 1.1\% to 1.6\% on the following word for GPT-2 on OneStop, and lengthens reading time by 0.2\% to 0.3\% on each for Pythia 410M to 2.8B on Natural Stories.
Pooled, the gain from $D$ is 2.1\% of the gain from final-layer surprisal of $w$ and $w-1$ over the baseline (0.26 against 12.05 millinats per word) and 30\% of the gain from final-layer entropy of $w$ and $w-1$ beyond the other controls (0.85).

\subsection{Layer order and untrained networks}
\label{sec:controls}

Shuffling the readouts removes the gain.
Under the five layer permutations, the pooled gain on the primary measure is $-0.016$ millinats per word ($-0.051$ to $0.021$; $p = 0.79$), and no model keeps a significant gain (Figure~\ref{fig:main}).
On total reading time the shuffled gain is $-0.01$ ($p = 0.66$); OPT 1.3B is the one model that keeps a significant shuffled gain there, 0.10 of its 0.57 ($p = 0.019$).
No untrained network reproduces the gain: on the primary measure, GPT-2 XL and OPT 1.3B with random weights gain $-0.07$ and 0.08 millinats per word and Pythia 410M and 1.4B at initialization $-0.05$ and $-0.09$ ($p \ge 0.27$), and the four together gain $-0.03$ ($p = 0.75$), against 0.22 ($p = 0.025$) for trained models read the same way, with logit-lens controls.
The untrained networks thus rule out architecture and tokenization as the source of the logit-lens effect, which lies mostly in the first step after the embedding readout (\S\ref{sec:robust}).
Three of the 104 untrained cells are positive and significant, all in GECO, against 5.2 expected by chance (Table~\ref{tab:cells-untrained}).

\subsection{Robustness and alternative measures}
\label{sec:robust}

\begin{table}[t]
\centering
\footnotesize
\setlength{\tabcolsep}{4pt}
\renewcommand{\arraystretch}{0.92}
\begin{tabular}{lrr}
\toprule
Analysis & Gain & $p$ \\
\midrule
Main test & 0.257 & 0.001 \\
\quad $D(w)$ alone & 0.094 & 0.015 \\
\quad $D(w-1)$ alone & 0.174 & 0.002 \\
\midrule
From readout 1 & 0.188 & 0.006 \\
Corrected word probabilities & 0.185 & 0.009 \\
Single-token words & 0.171 & 0.012 \\
+ function-word status & 0.258 & 0.001 \\
+ nonlinear length, frequency & 0.217 & 0.004 \\
+ JS and KL, first vs.\ last & 0.241 & 0.001 \\
+ quadratic $s_L$, SU & 0.187 & 0.004 \\
+ TV and SU of $w-1$ & 0.124 & 0.029 \\
+ skipping rate & 0.201 & 0.005 \\
$D$ of the entropy trajectory & $-0.017$ & 0.60 \\
\midrule
Logit lens, main controls & 0.441 & $<$0.001 \\
\quad from readout 1 & 0.164 & 0.010 \\
Logit lens, logit controls & 0.225 & 0.025 \\
\quad from readout 1 & 0.013 & 0.39 \\
\midrule
Mean layerwise surprisal alone & 0.137 & 0.003 \\
\quad $D$ beyond it & 0.101 & 0.067 \\
\quad $D$ beyond its quadratic terms & 0.070 & 0.15 \\
Prediction depth alone & $-0.004$ & 0.52 \\
\quad $D$ beyond it & 0.206 & 0.003 \\
Largest-drop depth alone & 0.014 & 0.38 \\
\quad $D$ beyond it & 0.126 & 0.019 \\
Settling depth alone & $-0.013$ & 0.63 \\
\quad $D$ beyond it & 0.165 & 0.007 \\
\bottomrule
\end{tabular}
\caption{Robustness and alternative measures: pooled held-out gain (millinats per word) on the primary measure over eight corpora and ten models, with one-sided sign-flip $p$. ``+'' rows add the named variables of $w$ and $w-1$ to the controls before testing $D$; ``alone'' rows test the alternative measure in place of $D$, and ``beyond'' rows test $D$ with the alternative in the controls.}
\label{tab:robust}
\end{table}

Leaving out the embedding readout (0.19 millinats per word), correcting word probabilities for word boundaries (0.19) and restricting the test to single-token words (0.17) keep the gain ($p \le 0.012$; Table~\ref{tab:robust}), so the embedding readout and token segmentation do not produce it.
With function-word status, nonlinear terms in length and frequency, the skipping rates of $w$ and $w-1$, the first-to-last-layer divergences of \citet{kuribayashi2026dual} or quadratic terms in the trajectory's endpoints in the controls, the gain stays at 0.19 to 0.26 ($p \le 0.005$).
With the total variation and surprisal update of $w-1$ also in the controls, $D$ gains 0.12 ($p = 0.029$; OPT 0.17, $p = 0.035$; GPT-2 and Pythia 0.10, $p = 0.12$ and $0.10$) and 0.14 on total reading time ($p = 0.014$): part of $D(w-1)$'s information is the size of the preceding word's change.
$D$ of the entropy trajectory adds nothing on the primary measure ($-0.02$; $p = 0.60$) and predicts total reading time nearly as well as $D$ (0.30 against 0.34); on the primary measure, the information is specific to the surprisal trajectory of the word that occurs.

The logit lens is a coarser instrument \citep{belrose2023eliciting}: logit-lens $D$ gains 0.44 millinats per word beyond the main controls ($p < 0.001$), much of it from the first step of GPT-2 and OPT, whose tied embeddings make the first readout assign tens of nats to the next word, and 0.16 from readout 1 ($p = 0.010$); with logit-lens controls it gains 0.22 ($p = 0.025$), and 0.01 from readout 1 ($p = 0.39$).

The mean surprisal of the trajectory, unchanged when the intermediate readouts are reordered, predicts reading times alone (0.14; $p = 0.003$); for a decreasing trajectory it equals $s_L(w) + D(w)\,\mathrm{SU}(w)$ and carries the information of $D$ given the endpoints, and $D$ adds 0.10 beyond it ($p = 0.067$) and 0.07 beyond its quadratic terms with the final surprisal and update ($p = 0.15$; 0.12 on total reading time, $p = 0.033$).
Prediction depth, the largest-drop depth and the settling depth gain $-0.01$ to 0.01 alone ($p \ge 0.38$), and $D$ adds beyond each (0.21, 0.13 and 0.17; $p \le 0.019$): a centroid over the whole trajectory captures timing information that single landmarks miss.

\subsection{Functional regression on the whole trajectory}
\label{sec:res-functional}

\begin{table}[t]
\centering
\small
\setlength{\tabcolsep}{4pt}
\begin{tabular}{lrrr}
\toprule
Comparison & Gain & 95\% CI & $p$ \\
\midrule
$D$ & 0.257 & 0.10, 0.40 & 0.001 \\
Trajectory & 0.168 & $-$0.00, 0.34 & 0.047 \\
Trajectory beyond $D$ & $-0.020$ & $-$0.16, 0.12 & 0.61 \\
$D$ beyond trajectory & 0.069 & $-$0.05, 0.19 & 0.15 \\
Profile beyond $D$ & $-0.010$ & $-$0.12, 0.09 & 0.57 \\
\bottomrule
\end{tabular}
\caption{Functional regression on the trajectories of $w$ and $w-1$: pooled held-out gain (millinats per word) on the primary measure; ``beyond'' rows add one term to a model that contains the other (shuffled-layer rows in Table~\ref{tab:functional}).}
\label{tab:functional-main}
\end{table}

The functional regression on the whole trajectory gains 0.17 millinats per word ($p = 0.047$; Table~\ref{tab:functional-main}), less than $D$.
The two add little to each other in either direction, and the regression on the resolution profile, which reduces to $D$ when its coefficient function is linear, adds nothing beyond $D$.
The functional term gains little from the true layer order (0.05 over shuffled layers; $p = 0.14$) because shuffled trajectories keep the order-free mean surprisal (gain 0.12; $p = 0.048$), which adds $-0.05$ beyond $D$ ($p = 0.80$; Table~\ref{tab:functional} in Appendix~\ref{app:functional}).
Across corpora, the trajectory adds information beyond $D$ in OneStop (0.41; $p = 0.003$), and in Natural Stories it fits worse than $D$ while the shape of the resolution profile adds to $D$ (0.87; $p = 0.016$; Table~\ref{tab:functional-corpus} in Appendix~\ref{app:functional}).

\subsection{Which words resolve late}
\label{sec:words}

Comparing, within each decile of final surprisal, the fifth of words with the highest $D$ with the fifth with the lowest, late-resolved words are shorter in 74 of the 80 corpus--model pairs (median 0.44 letters), more frequent in 65 and more often function words in 60 (median share 55\% against 50\%).
In OneStop under GPT-2 XL, the commonest late-resolved words of middling surprisal are \textit{and}, \textit{in}, \textit{a}, \textit{the}, \textit{to}, \textit{for}, \textit{more} and \textit{of}, and the commonest early-resolved ones \textit{the}, \textit{have}, \textit{has}, \textit{is} and \textit{was}.
$D$ correlates positively with final surprisal in 79 of the 80 pairs (Spearman $-0.10$ to $0.44$), a dependence the controls absorb, and its median changes little with model size (GPT-2 small to XL 0.41 to 0.38, Pythia 410M to 6.9B 0.45 to 0.41, OPT 125M to 6.7B 0.49 to 0.48).

\section{Discussion}
\label{sec:discussion}

Pooled over corpora and models, where a language model resolves its prediction of a word along its depth predicts how long people read the word and the next one, beyond how predictable the word is.

A high $D$ means that the word's surprisal stays high through more of the network: the model completes its prediction late.
Surprisal from earlier readouts correlates more strongly with bigram surprisal, and surprisal from later readouts with fully contextualized surprisal \citep[\S7.1]{kuribayashi2025large}, so a late-resolved word is one that the local context leaves open and the wider context settles.
At matched surprisal, such words are shorter and more frequent, and in OneStop under GPT-2 XL the commonest of them are conjunctions, prepositions and determiners that begin a phrase; readers spend extra time on them and on the word that follows, which carries most of the gain (\S\ref{sec:main}).
\citet{barenholtz2026trajectory} found such connectives enriched among predictable words whose hidden states depart from the path extrapolated from the preceding words; both measures mark where a sentence's structure moves on.
Function-word status as a control leaves the gain unchanged (\S\ref{sec:robust}), and \textit{the} is common among both late- and early-resolved words: $D$ separates occurrences of the same word.

Under surprisal theory, a word's processing difficulty depends on the comprehender's representations only through its conditional probability, the causal bottleneck of \citet{levy2008expectationbased}.
$D$ predicts reading times beyond final-layer surprisal, entropy and the best single readout, and two accounts fit this result.
In the first, the probability that governs human reading differs from every individual readout, and the shape of the trajectory refines the estimate.
In the second, the cost of a word includes a component that grows with the computation its prediction requires, in the spirit of depth-of-computation measures of example difficulty \citep{baldock2021deep}.

The result extends the layer-wise analyses of \citet{kuribayashi2025large} and the first-to-last-layer measures of \citet{kuribayashi2026dual}: $D$ adds to the best readout selected within each training fold and to the size of the change between two layers.
Flexible functions of the trajectory's endpoints leave $D$'s gain in place, and the mean surprisal of the trajectory recovers most of it, so the information lies in the intermediate readouts; pooled over corpora, a functional regression on all of them adds no significant information beyond $D$, and OneStop and Natural Stories hold more in the trajectory's shape (\S\ref{sec:res-functional}).
That mean is the word's surprisal under a weighted geometric mean of the readouts' probabilities, an estimate of the kind the first account posits; for a decreasing trajectory it carries $D$'s information given the endpoints, and $D$ fits reading times better.

The net effect depends on corpus and model family (\S\ref{sec:main}); OPT gains in OneStop (0.35; $p = 0.023$), GECO (0.14; $p = 0.043$) and Natural Stories (0.73; $p = 0.055$).
For GPT-2 on Natural Stories the full-data coefficients of $D(w)$ and $D(w-1)$ are at most 0.08\% per standard deviation, so $D$ carries no signal there; its losses ($-0.28$ to $-0.39$) exceed the $1/n$ approximation (0.12) and lie within the range of shuffled and untrained $D$ on the same ten stories ($-0.39$ to $0.35$): with the words clustered in ten texts, coefficients without signal cost more than $1/n$.
GECO, the largest corpus, gives gains near zero although its reader count, surprisal and spread of $D$ do not set it apart (final surprisal 3.1 to 4.2 nats; standard deviation of $D$ 0.07 to 0.11); telling which properties of a text, a reader population or a model family govern the effect requires more corpora of each kind.

Pooled over corpora, $D$ is an inexpensive, interpretable complement to surprisal: it needs one forward pass and a lens and compares models of different depth.

\section{Limitations}
\label{sec:limitations}

\paragraph{Size and heterogeneity.}
The pooled gain is small and its net size depends on the corpus: for GPT-2 and Pythia it rests on one corpus each and loses significance with the total variation and surprisal update of $w-1$ in the controls (\S\ref{sec:robust}), the six corpora other than OneStop and Natural Stories give 0.06 ($p = 0.16$), and the prediction interval for a new corpus spans $-0.93$ to $1.07$ millinats per word ($-0.14$ to $0.60$ against layer-shuffled $D$; \S\ref{sec:main}).
The claims concern the pooled effect over corpora and models.

\paragraph{Lenses.}
We use the released tuned lenses; the 2025 version of the tuned-lens paper (arXiv v6; \citealp{belrose2023eliciting}) reports that lenses trained with the Muon optimizer, which orthogonalizes the update of each weight matrix, reach much lower divergence from the output, and describes the released lenses as severely undertrained.
Under the logit lens with logit-lens controls and without the embedding readout the gain is 0.01 ($p = 0.39$; Table~\ref{tab:robust}), so the effect is measured with learned translators, and its size under better-trained lenses remains open.
Tuned lenses fitted to the untrained networks would extend their control to the readouts after the embedding.

\paragraph{Training data.}
ZuCo, MECO and two Natural Stories texts derive from Wikipedia, which the training data of OPT and Pythia include and GPT-2's excludes (\S\ref{sec:data}); without those two stories, Pythia still gains 0.76 millinats per word on Natural Stories ($p = 0.036$) and GPT-2 $-0.53$.

\paragraph{Aggregation and scope.}
We model reading times averaged over readers, which removes reader-level variation and does not test individual differences, and compute word entropy over the first token, a lower bound on the word entropy \citep{pimentel2023effect}.
The study covers English reading by first-language readers and three families of decoder-only transformers up to 6.9B parameters, and its analyses are exploratory (\S\ref{sec:analysis}).
It reanalyzes published, anonymized reading-time corpora under their licenses and collects no new data from human participants.

\section{Conclusion}

Pooled over eight English reading corpora and ten models, words that a language model resolves later along its depth, and the words after them, are read more slowly than equally surprising words, and resolution depth carries this information beyond surprisal, entropy and the best single layer.
Experiments that vary the computation a prediction needs at a fixed probability would decide whether that computation itself costs human readers time.

\section*{Acknowledgments}
The authors used a generative AI system (Claude, Anthropic), under their supervision throughout, in writing the analysis code and drafting the manuscript text. The authors take full responsibility for the content of the paper.

\bibliography{references}
\bibliographystyle{acl_natbib}

\appendix

\section{Implementation}
\label{app:impl}

For Natural Stories we use the release in the corpus repository (\url{https://github.com/languageMIT/naturalstories}) at commit \texttt{4700daa} of 14 May 2025, which corrects a one-position misalignment between words and self-paced reading times.
Natural Stories contributes the corpus's processed reading times from readers who answered at least five of six comprehension questions for a story, and UCL the native English readers who answered at least 75\% of the comprehension questions correctly, the original criterion (62 self-paced and 37 eye-tracking readers); we keep self-paced reading times between 100 and 3,000\,ms.
We exclude the first and last word of each sentence and of each display line (sentence edges only where line positions are unavailable or a sentence fills one line: MECO, ZuCo, UCL and Natural Stories), words without letters, and words whose labels disagree between the text and the reading data.
Counting each word under the first criterion that applies, the primary analysis excludes, of the 118,134 displayed words, 26,815 sentence and line edges, 194 words without letters, 778 words whose labels disagree, 7,253 words with fewer contributing readers than the minimum and 1,137 words with an undefined predictor of $w-1$ or $w-2$ (the second word of a document, or a preceding word without a bigram surprisal).
By corpus, in this order: Provo 530, 1, 44, 1 and 54; GECO 16,219, 0, 0, 4,515 and 4; OneStop 7,123, 159, 734, 1,638 and 175; MECO 202, 2, 0, 157 and 14; ZuCo 600, 32, 0, 931 and 314; UCL eye tracking 410, 0, 0, 11 and 205; Natural Stories 1,009, 0, 0, 0 and 10; UCL self-paced 722, 0, 0, 0 and 361.
In UCL eye tracking the release provides first fixation, gaze and go-past times, and total reading time sums every fixation on the word in the release's fixation records.

Each Provo passage, part of the GECO novel (four parts), OneStop article version, MECO text and Natural Stories story is processed as one continuous document, and each ZuCo and UCL sentence, presented to readers on its own, as a separate document.
A document's words are joined by single spaces, preceded by the end-of-text token and read in windows of 1,024 tokens with a stride of 512, so that every token beyond the first window is predicted from at least 511 tokens of context.
Word surprisal sums the surprisals of the word's tokens, the first of which carries the preceding space.
This sum omits the probability that the word ends after its last token \citep{oh2024leading,pimentel2024compute}; the corrected surprisal at readout $i$ is $s_i(w) - \log B_i(c\,w) + \log B_i(c)$, where $B_i(x)$ is the probability that readout $i$ assigns, after $x$, to a token that begins with a space or ends the text \citep{pimentel2024compute}.
Word entropy is the entropy of the readout at the position that predicts the word's first token, a lower bound on the entropy of the word distribution \citep{pimentel2023effect}.
Tuned-lens translators are the released parameter files for each model, applied to the model revision each lens was trained on \citep{belrose2023eliciting}; computations use 32-bit floating point.
The untrained GPT-2 XL and OPT 1.3B use the default initialization of their Hugging Face implementations (normal weights with standard deviation 0.02) under random seed 0; the Pythia controls are the released step-0 checkpoints \citep{biderman2023pythia}.

Zipf frequencies come from wordfreq release 3.1.1, whose word lists equal those of the archived release 3.0.2 \citep{speer2022wordfreq}, and function words are those of the English stop-word list of scikit-learn \citep{pedregosa2011scikit}.
The functional predictors are centered and divided by one common scale, so that the roughness penalty, which acts on the functional terms alone with one weight for the terms of $w$ and $w-1$, acts on $\beta$ itself.
Regressions are solved by minimum-norm least squares (singular value decomposition) on the weighted, penalty-augmented design.
Some predictors are exactly collinear for every word (the readout-0 surprisal equals the surprisal update plus the final surprisal, and the spline terms of the resolution profile span the intercept and $D$), so held-out predictions are unique.
Held-out log-likelihood is the Gaussian log density of $y_w$ with variance $\hat\sigma^2 / n_w$, where $\hat\sigma^2$, the maximum-likelihood estimate in the training fold, is the mean over training words of $n_w$ times the squared residual.
In the random-effects models each corpus enters with the standard error of its gain over 4,000 bootstrap samples of its text units; counted as one corpus, the two UCL corpora sum each sentence's gains over both and divide by their combined word predictions.
The Hartung--Knapp variance factor is truncated below at one; the Hartung--Knapp interval and the one-sided test use a $t$ distribution with $k-1$ degrees of freedom for $k$ corpora, and the prediction interval one with $k-2$.
The reader minimum, the eight B-splines, the 10\% settling threshold, the five layer permutations and the random seed of the untrained networks are fixed settings that we did not tune; each permutation reruns the complete analysis, which limits their number to five, and each untrained network is a single initialization.

\twocolumn[{%
\section{Results for Every Corpus, Model and Measure}
\label{app:cells}
\begin{center}
\footnotesize
\setlength{\tabcolsep}{4.5pt}
\renewcommand{\arraystretch}{0.94}
\begin{tabular}{lrrrr}
\toprule
Model & FFD & GD & GP & TRT \\
\midrule
\multicolumn{5}{l}{\textit{Provo}} \\
GPT-2 small & $-0.99$ & $-0.73$ & $-0.07$ & $-0.29$ \\
GPT-2 large & ${+}0.61$ & ${+}0.96$ & $-0.42$ & $-0.09$ \\
GPT-2 XL & ${+}1.98^{*}$ & ${+}1.07$ & ${+}1.03$ & ${+}0.85$ \\
OPT 125M & ${+}0.72$ & ${+}1.09$ & ${+}0.72$ & $-0.41$ \\
OPT 1.3B & $-0.01$ & $-0.20$ & $-0.28$ & $-0.47$ \\
OPT 6.7B & ${+}0.22$ & $-0.44$ & ${+}0.66$ & $-0.18$ \\
Pythia 410M & $-0.54$ & $-1.19^{\dagger}$ & $-0.42$ & $-1.64^{\dagger}$ \\
Pythia 1.4B & ${+}0.03$ & ${+}0.13$ & $-1.01^{\dagger}$ & $-0.84^{\dagger}$ \\
Pythia 2.8B & ${+}0.12$ & $-1.02^{\dagger}$ & $-0.89^{\dagger}$ & $-0.81^{\dagger}$ \\
Pythia 6.9B & $-0.68$ & $-1.17^{\dagger}$ & $-0.52$ & $-1.29$ \\
\addlinespace
\multicolumn{5}{l}{\textit{GECO}} \\
GPT-2 small & $-0.08^{\dagger}$ & $-0.08$ & $0.00$ & $-0.02$ \\
GPT-2 large & $0.00$ & $-0.08^{\dagger}$ & $0.00$ & $-0.03$ \\
GPT-2 XL & $0.00$ & ${+}0.09$ & ${+}0.17$ & ${+}0.22^{*}$ \\
OPT 125M & $-0.04$ & $-0.04$ & ${+}0.02$ & ${+}0.05$ \\
OPT 1.3B & ${+}0.06$ & $0.00$ & ${+}0.09$ & ${+}0.15$ \\
OPT 6.7B & ${+}0.09$ & $-0.02$ & ${+}0.31^{*}$ & $-0.02$ \\
Pythia 410M & $-0.06$ & $-0.08$ & $-0.03$ & $0.00$ \\
Pythia 1.4B & ${+}0.01$ & $-0.02$ & ${+}0.03$ & ${+}0.01$ \\
Pythia 2.8B & $-0.05^{\dagger}$ & $-0.04$ & ${+}0.05$ & ${+}0.07$ \\
Pythia 6.9B & $-0.06$ & $-0.05$ & ${+}0.12$ & ${+}0.02$ \\
\addlinespace
\multicolumn{5}{l}{\textit{OneStop}} \\
GPT-2 small & ${+}0.26$ & ${+}0.42$ & ${+}0.78^{*}$ & ${+}0.88^{*}$ \\
GPT-2 large & ${+}0.49^{*}$ & ${+}0.82^{*}$ & ${+}1.11^{*}$ & ${+}0.79$ \\
GPT-2 XL & ${+}1.02^{*}$ & ${+}2.59^{*}$ & ${+}1.47^{*}$ & ${+}1.99^{*}$ \\
OPT 125M & ${+}0.05$ & ${+}0.39^{*}$ & ${+}0.49^{*}$ & ${+}0.12$ \\
OPT 1.3B & ${+}0.11$ & ${+}0.70^{*}$ & ${+}0.30$ & ${+}1.03^{*}$ \\
OPT 6.7B & ${+}0.48^{*}$ & ${+}0.70^{*}$ & ${+}0.26$ & ${+}1.02^{*}$ \\
Pythia 410M & ${+}0.23$ & ${+}0.44$ & ${+}0.24$ & ${+}0.55^{*}$ \\
Pythia 1.4B & ${+}0.16$ & ${+}0.43^{*}$ & ${+}0.17$ & ${+}0.28$ \\
Pythia 2.8B & ${+}0.06$ & ${+}0.06$ & ${+}0.01$ & ${+}0.25$ \\
Pythia 6.9B & ${+}0.11$ & ${+}0.20$ & ${+}0.26$ & ${+}0.53^{*}$ \\
\addlinespace
\multicolumn{5}{l}{\textit{Natural Stories}} \\
GPT-2 small & -- & -- & $-0.33^{\dagger}$ & -- \\
GPT-2 large & -- & -- & $-0.28$ & -- \\
GPT-2 XL & -- & -- & $-0.39$ & -- \\
OPT 125M & -- & -- & $-0.26^{\dagger}$ & -- \\
OPT 1.3B & -- & -- & ${+}0.79$ & -- \\
OPT 6.7B & -- & -- & ${+}1.65^{*}$ & -- \\
Pythia 410M & -- & -- & ${+}1.17^{*}$ & -- \\
Pythia 1.4B & -- & -- & ${+}1.64^{*}$ & -- \\
Pythia 2.8B & -- & -- & ${+}1.50^{*}$ & -- \\
Pythia 6.9B & -- & -- & ${+}0.01$ & -- \\
\bottomrule
\end{tabular}
\hspace{2.5em}
\begin{tabular}{lrrrr}
\toprule
Model & FFD & GD & GP & TRT \\
\midrule
\multicolumn{5}{l}{\textit{MECO}} \\
GPT-2 small & ${+}1.00$ & ${+}3.90$ & ${+}0.93$ & ${+}1.57$ \\
GPT-2 large & $-0.85$ & $-0.42$ & $-0.86$ & $-2.49$ \\
GPT-2 XL & $-0.40$ & ${+}1.03$ & $-0.62$ & $-0.18$ \\
OPT 125M & ${+}0.61$ & ${+}3.20$ & $-0.49$ & ${+}1.11$ \\
OPT 1.3B & ${+}1.09$ & ${+}2.71$ & ${+}0.72$ & ${+}3.01^{*}$ \\
OPT 6.7B & ${+}0.83$ & ${+}1.78$ & $-0.62$ & ${+}0.77$ \\
Pythia 410M & $-0.86$ & $-0.70$ & $-0.38$ & $-2.02$ \\
Pythia 1.4B & $-0.80$ & $-0.80$ & $-0.62$ & $-1.67^{\dagger}$ \\
Pythia 2.8B & $-0.75$ & $-0.67$ & ${+}0.98$ & ${+}0.26$ \\
Pythia 6.9B & $-0.31$ & ${+}0.56$ & $-1.03$ & $-0.51$ \\
\addlinespace
\multicolumn{5}{l}{\textit{ZuCo}} \\
GPT-2 small & $0.00$ & ${+}0.29$ & ${+}1.20$ & ${+}0.81$ \\
GPT-2 large & ${+}0.39$ & ${+}0.23$ & ${+}0.01$ & ${+}0.03$ \\
GPT-2 XL & ${+}1.36^{*}$ & ${+}2.21^{*}$ & ${+}0.55$ & ${+}0.98$ \\
OPT 125M & $-0.24$ & $-0.17$ & $-0.25$ & ${+}0.04$ \\
OPT 1.3B & ${+}0.22$ & ${+}1.28$ & $-0.07$ & ${+}0.86$ \\
OPT 6.7B & $-0.03$ & ${+}1.27$ & ${+}0.56$ & ${+}0.83$ \\
Pythia 410M & $-0.30$ & ${+}0.79$ & $-0.27$ & ${+}0.56$ \\
Pythia 1.4B & ${+}0.39$ & ${+}1.04$ & ${+}0.10$ & ${+}0.42$ \\
Pythia 2.8B & ${+}0.55$ & ${+}1.30$ & ${+}0.95$ & ${+}2.11^{*}$ \\
Pythia 6.9B & ${+}0.07$ & ${+}0.95$ & ${+}0.19$ & ${+}0.96$ \\
\addlinespace
\multicolumn{5}{l}{\textit{UCL eye tracking}} \\
GPT-2 small & $-1.68^{\dagger}$ & $-2.25^{\dagger}$ & $-1.37$ & $-0.76$ \\
GPT-2 large & $-0.68$ & $-0.80$ & ${+}0.04$ & $-1.62^{\dagger}$ \\
GPT-2 XL & $-2.41^{\dagger}$ & $-1.96^{\dagger}$ & $-2.84^{\dagger}$ & $-0.79$ \\
OPT 125M & $-1.39^{\dagger}$ & $-1.15$ & $-2.02^{\dagger}$ & $-1.30$ \\
OPT 1.3B & $-1.38$ & $-1.02$ & $-0.80$ & $-1.34^{\dagger}$ \\
OPT 6.7B & $-2.91^{\dagger}$ & $-2.86^{\dagger}$ & $-1.68$ & $-0.58$ \\
Pythia 410M & $-1.11$ & ${+}0.72$ & $-0.12$ & $-0.82$ \\
Pythia 1.4B & $-0.18$ & ${+}0.27$ & $-0.70$ & $-3.97^{\dagger}$ \\
Pythia 2.8B & $-1.32$ & ${+}0.21$ & $-1.79^{\dagger}$ & ${+}2.10$ \\
Pythia 6.9B & $-2.26^{\dagger}$ & $-1.49^{\dagger}$ & $-1.67^{\dagger}$ & $-0.11$ \\
\addlinespace
\multicolumn{5}{l}{\textit{UCL self-paced}} \\
GPT-2 small & -- & -- & $-0.38$ & -- \\
GPT-2 large & -- & -- & ${+}1.09$ & -- \\
GPT-2 XL & -- & -- & ${+}0.19$ & -- \\
OPT 125M & -- & -- & ${+}0.29$ & -- \\
OPT 1.3B & -- & -- & ${+}0.52$ & -- \\
OPT 6.7B & -- & -- & ${+}0.29$ & -- \\
Pythia 410M & -- & -- & $-0.43$ & -- \\
Pythia 1.4B & -- & -- & ${+}0.96$ & -- \\
Pythia 2.8B & -- & -- & ${+}0.48$ & -- \\
Pythia 6.9B & -- & -- & ${+}0.61$ & -- \\
\bottomrule
\end{tabular}
\tablecaption{Held-out gain (millinats per word) from adding $D(w)$ and $D(w-1)$ to the control set, for each corpus, trained model and measure: first fixation duration (FFD), gaze duration (GD), go-past time (GP) and total reading time (TRT). The self-paced reading corpora have one measure, the reading time, listed under GP, so the table holds 260 cells. Asterisks mark one-sided permutation $p < 0.05$ for the cell alone (10,000 sign flips over text units), daggers a significant loss (two-sided $p < 0.05$).}
\label{tab:cells}
\end{center}
}]

\twocolumn[{%
\renewcommand{\arraystretch}{0.92}
\begin{center}
\footnotesize
\setlength{\tabcolsep}{3.5pt}
\begin{tabular}{lrrrr}
\toprule
Model & FFD & GD & GP & TRT \\
\midrule
\multicolumn{5}{l}{\textit{Provo}} \\
GPT-2 XL, random & $-0.62$ & $-0.76^{\dagger}$ & $-0.20$ & $-0.31$ \\
OPT 1.3B, random & $-1.16$ & $-1.59^{\dagger}$ & $-0.19$ & $-0.89$ \\
Pythia 410M, step 0 & $-0.84$ & $-0.96$ & $-1.07^{\dagger}$ & $-1.00$ \\
Pythia 1.4B, step 0 & $-1.22$ & ${+}0.16$ & $-1.00^{\dagger}$ & $-0.68$ \\
\addlinespace
\multicolumn{5}{l}{\textit{GECO}} \\
GPT-2 XL, random & $-0.05^{\dagger}$ & $-0.02$ & $-0.03$ & ${+}0.01$ \\
OPT 1.3B, random & ${+}0.04$ & ${+}0.28^{*}$ & ${+}0.09$ & ${+}0.36^{*}$ \\
Pythia 410M, step 0 & $-0.05$ & $-0.04$ & ${+}0.04$ & $-0.02$ \\
Pythia 1.4B, step 0 & ${+}0.04$ & ${+}0.05$ & ${+}0.06$ & ${+}0.34^{*}$ \\
\addlinespace
\multicolumn{5}{l}{\textit{OneStop}} \\
GPT-2 XL, random & $-0.11$ & $-0.07$ & ${+}0.07$ & $-0.07$ \\
OPT 1.3B, random & $-0.20^{\dagger}$ & $-0.15$ & $-0.06$ & $-0.20^{\dagger}$ \\
Pythia 410M, step 0 & $0.00$ & $-0.09$ & $-0.06$ & $-0.07$ \\
Pythia 1.4B, step 0 & $-0.06$ & $-0.06$ & $-0.03$ & ${+}0.05$ \\
\addlinespace
\multicolumn{5}{l}{\textit{Natural Stories}} \\
GPT-2 XL, random & -- & -- & $-0.06$ & -- \\
OPT 1.3B, random & -- & -- & $-0.04$ & -- \\
Pythia 410M, step 0 & -- & -- & ${+}0.17$ & -- \\
Pythia 1.4B, step 0 & -- & -- & $-0.39$ & -- \\
\bottomrule
\end{tabular}
\hspace{1em}
\begin{tabular}{lrrrr}
\toprule
Model & FFD & GD & GP & TRT \\
\midrule
\multicolumn{5}{l}{\textit{MECO}} \\
GPT-2 XL, random & $-0.85$ & $-1.15$ & $-1.73^{\dagger}$ & $-1.82$ \\
OPT 1.3B, random & $-0.86$ & ${+}0.15$ & ${+}2.61$ & ${+}0.23$ \\
Pythia 410M, step 0 & ${+}2.90$ & ${+}1.67$ & ${+}0.53$ & ${+}0.86$ \\
Pythia 1.4B, step 0 & $-0.48$ & ${+}1.05$ & ${+}0.21$ & ${+}0.55$ \\
\addlinespace
\multicolumn{5}{l}{\textit{ZuCo}} \\
GPT-2 XL, random & $-0.06$ & $-0.38$ & $-0.03$ & $-0.78^{\dagger}$ \\
OPT 1.3B, random & ${+}0.45$ & ${+}0.12$ & ${+}0.94$ & $-0.25$ \\
Pythia 410M, step 0 & ${+}0.07$ & $-0.37$ & $-0.21$ & $-0.49$ \\
Pythia 1.4B, step 0 & $-0.05$ & $-0.71^{\dagger}$ & $-0.21$ & $-0.24$ \\
\addlinespace
\multicolumn{5}{l}{\textit{UCL eye tracking}} \\
GPT-2 XL, random & ${+}0.78$ & $-0.52$ & $-0.95$ & $-1.23$ \\
OPT 1.3B, random & $-0.64$ & $-0.53$ & $-0.26$ & ${+}0.57$ \\
Pythia 410M, step 0 & $-0.11$ & $-1.43$ & $-1.27$ & ${+}0.10$ \\
Pythia 1.4B, step 0 & $-2.57^{\dagger}$ & $-3.04^{\dagger}$ & $-2.25$ & $-1.61^{\dagger}$ \\
\addlinespace
\multicolumn{5}{l}{\textit{UCL self-paced}} \\
GPT-2 XL, random & -- & -- & $-0.25$ & -- \\
OPT 1.3B, random & -- & -- & $-0.86^{\dagger}$ & -- \\
Pythia 410M, step 0 & -- & -- & $-0.23$ & -- \\
Pythia 1.4B, step 0 & -- & -- & ${+}0.03$ & -- \\
\bottomrule
\end{tabular}
\tablecaption{Held-out gain (millinats per word) from adding $D(w)$ and $D(w-1)$ to the control set for the untrained networks, read with the logit lens: GPT-2 XL and OPT 1.3B with random weights and Pythia 410M and 1.4B at initialization. Columns, asterisks and daggers as in Table~\ref{tab:cells}.}
\label{tab:cells-untrained}
\end{center}
\begin{center}
\includegraphics[width=0.48\textwidth]{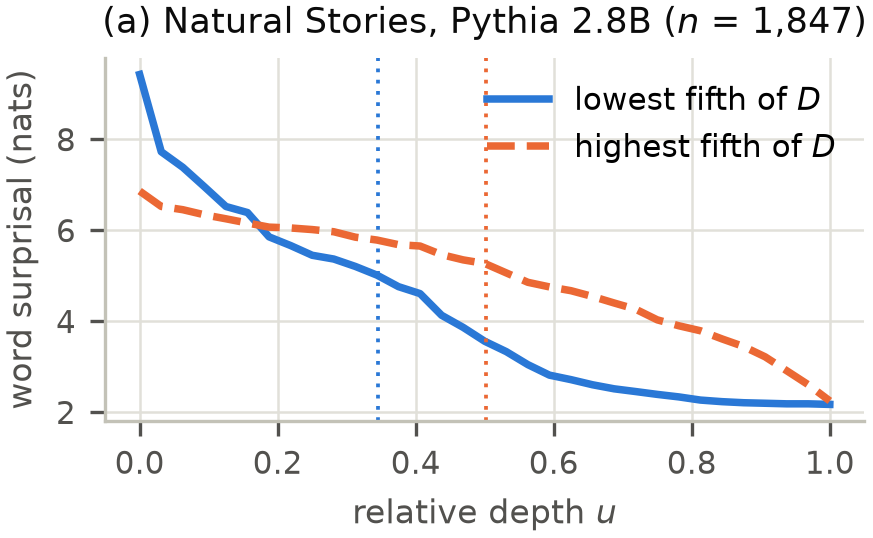}\hfill
\includegraphics[width=0.48\textwidth]{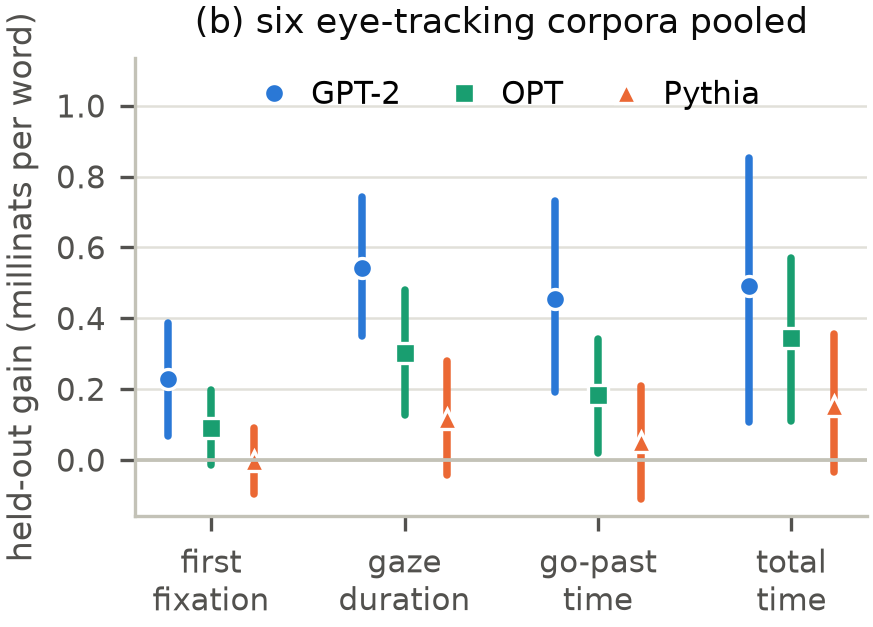}
\figurecaption{(a) Mean layerwise surprisal trajectories of Natural Stories words under Pythia 2.8B, as in Figure~\ref{fig:trajectories}. (b) Held-out gain from adding $D(w)$ and $D(w-1)$ to the control set, the six eye-tracking corpora pooled per measure and family, with 95\% bootstrap intervals over text units drawn within corpora.}
\label{fig:appendix}
\end{center}
}]

\twocolumn[{%
\section{Functional Regression}
\label{app:functional}
\begin{center}
\small
\setlength{\tabcolsep}{5pt}
\begin{tabular}{lrrr}
\toprule
Comparison & Gain & 95\% CI & $p$ \\
\midrule
$D$ & 0.257 & 0.10, 0.40 & 0.001 \\
Trajectory & 0.168 & $-$0.00, 0.34 & 0.047 \\
\quad true vs.\ shuffled & 0.047 & $-$0.03, 0.12 & 0.14 \\
Trajectory beyond $D$ & $-0.020$ & $-$0.16, 0.12 & 0.61 \\
$D$ beyond trajectory & 0.069 & $-$0.05, 0.19 & 0.15 \\
Profile beyond $D$ & $-0.010$ & $-$0.12, 0.09 & 0.57 \\
Shuffled trajectory & 0.121 & 0.00, 0.25 & 0.048 \\
\quad beyond $D$ & $-0.047$ & $-$0.15, 0.05 & 0.80 \\
\bottomrule
\end{tabular}
\tablecaption{Functional regression on the layerwise surprisal trajectory of $w$ and $w-1$: pooled held-out gain (millinats per word) on the primary measure over eight corpora and ten models. The first two rows add $D$ or the trajectory term to the control set; ``beyond'' rows add one term to a model that already contains the other. ``Profile beyond $D$'' adds a functional term on the resolution profile, whose mean is $D$; shuffled rows use the median over five layer permutations.}
\label{tab:functional}
\end{center}
\begin{center}
\small
\begin{tabular}{lrrrrr}
\toprule
Corpus & $D$ & Trajectory & Trajectory beyond $D$ & $D$ beyond trajectory & Profile beyond $D$ \\
\midrule
Provo & $-0.12$ & ${+}0.35$ & $-0.08$ & $-0.55^{\dagger}$ & $-1.09$ \\
GECO & ${+}0.08$ & ${+}0.13^{*}$ & ${+}0.09$ & ${+}0.03$ & $-0.05$ \\
OneStop & ${+}0.51^{*}$ & ${+}0.75^{*}$ & ${+}0.41^{*}$ & ${+}0.16$ & ${+}0.06$ \\
Natural Stories & ${+}0.55^{*}$ & $-0.43$ & $-0.57$ & ${+}0.41$ & ${+}0.87^{*}$ \\
MECO & $-0.20$ & $-1.16$ & $-1.41$ & $-0.45$ & $-1.06$ \\
ZuCo & ${+}0.30$ & $-0.28$ & $-0.65^{\dagger}$ & $-0.08$ & $-0.37$ \\
UCL eye tracking & $-1.30^{\dagger}$ & $-1.88^{\dagger}$ & $-1.94^{\dagger}$ & $-1.36^{\dagger}$ & $-0.45$ \\
UCL self-paced & ${+}0.36$ & $-0.15$ & $-0.35$ & ${+}0.16$ & $-0.53$ \\
\bottomrule
\end{tabular}
\tablecaption{Functional regression per corpus on the primary measure, pooled over the ten trained models (millinats per word): $D$ and the whole-trajectory term over the control set, each beyond the other, and the resolution-profile term beyond $D$. Asterisks mark one-sided sign-flip $p < 0.05$ over the corpus's text units, daggers a significant loss (two-sided $p < 0.05$).}
\label{tab:functional-corpus}
\end{center}
}]

\end{document}